\documentclass{article}
\usepackage{ijcai24}

\usepackage{times}
\usepackage{soul}
\usepackage{url}
\usepackage[hidelinks]{hyperref}
\usepackage[utf8]{inputenc}
\usepackage[small]{caption}
\usepackage{graphicx}
\usepackage{amsmath}
\usepackage{amsthm}
\usepackage{booktabs}
\usepackage{algorithm}
\usepackage{algorithmic}
\usepackage[switch]{lineno}

\usepackage{times}
\usepackage{epsfig}
\usepackage{graphicx}
\usepackage{amsmath}
\usepackage{amssymb}
\usepackage{booktabs}
\usepackage{float}
\usepackage{xcolor}
\usepackage{multirow}

\usepackage[utf8]{inputenc}
\usepackage{pgfplots}
\DeclareUnicodeCharacter{2212}{−}
\usepgfplotslibrary{groupplots,dateplot}
\usetikzlibrary{patterns,shapes.arrows}
\pgfplotsset{compat=newest}

\title{ToDo: Token Downsampling for Efficient Generation of High-Resolution Images }

\author{
Ethan Smith$^1$
\and
Nayan Saxena$^1$\and
Aninda Saha$^1$\\
\affiliations
$^1$Leonardo AI Research Lab, North Sydney, NSW, Australia \\
\emails
\{ethan, nayan.saxena, aninda\}@leonardo.ai
}
\begin{document}

\maketitle

\begin{abstract}
   % Attention has been a key factor in the improving quality of Image Diffusion Models~\cite{DDPM}, but with a computational complexity that grows quadratically with sequence length, we are often restricted in the size of images that we can process within both reasonable time and plausible memory budgets. In this paper we explore the importance of global attention, specifically in the context of generative image models, which we argue have features with high redundancy in their informational content, making them a prime candidate for sparser attention methods. We propose a training-free method that merges or subsamples the key and value tokens used in attention to accelerate Stable Diffusion inference up to 2x for common sizes and up to 4x and beyond for high resolutions like 1536x1536 and beyond. We compare our method to a previous method of token merging~\cite{tomesd} and show both increased speed and fidelity. We also show that with the reduction in memory and computation time, as well as the out of the box differentiability of our method, we are capable of finetuning Stable Diffusion at very large sizes previously not possible. 

    Attention has been a crucial component in the success of image diffusion models, however, their quadratic computational complexity limits the sizes of images we can process within reasonable time and memory constraints. This paper investigates the importance of dense attention in generative image models, which often contain redundant features, making them suitable for sparser attention mechanisms. We propose a novel training-free method ToDo that relies on token downsampling of key and value tokens to accelerate Stable Diffusion inference by up to 2x for common sizes and up to 4.5x or more for high resolutions like $2048\times2048$. We demonstrate that our approach outperforms previous methods in balancing efficient throughput and fidelity.

    % Compared to previous token merging methods~\cite{tomesd}, our approach exhibits increased speed and fidelity 
    
    % Additionally, our method's differentiablity enables fine-tuning Stable Diffusion at previously unattainable large sizes, owing to reduced memory and computation requirements.
   
\end{abstract}

%------------------------------------------------------------------------- 
\section{Introduction}

Transformers, and their key component, attention, have been integral to the success and improvements in generative models in recent years~\cite{vaswani2023attention}. Their global receptive fields, ability to compute dynamically based on input context, and large capacities have made them useful building blocks across many tasks \cite{Khan_2022}. The main drawback of Transformer architectures is their quadratic scaling of computational complexity with sequence length, affecting both time and memory requirements. When looking to generate a Stable Diffusion image at $2048\times2048$  resolution, the attention maps of the largest U-Net blocks incur a memory cost of approximately 69~GB in half-precision, calculated as $(1 \text{ batch} \times 8 \text{ heads} \times (256^2 \text{ tokens})^2 \times 2 \text{ bytes})$. This exceeds the capabilities of most consumer GPUs~\cite{zhuang2023survey}. Specialized kernels, such as those used in Flash Attention, have greatly improved speed and reduced memory costs~\cite{dao2022flashattention}, however, challenges due to the unfavorable quadratic scaling with sequence length are persistent.

\begin{figure}[htbp]
\centering
\includegraphics[width= \columnwidth]{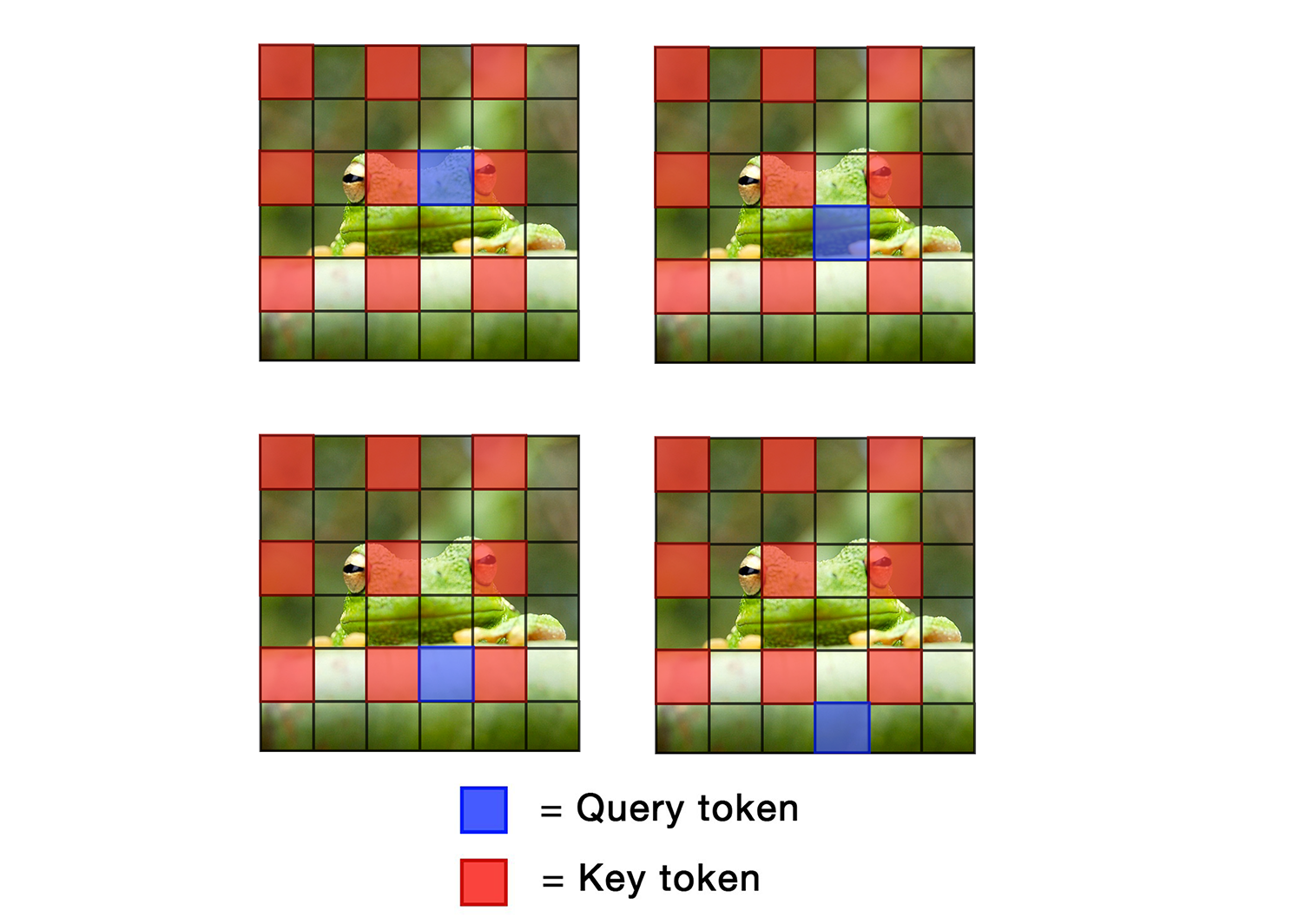}
\caption{{A visualization of our method. From a given latent or image, we subsample positions on the grid in a strided fashion for the keys and values used in attention maintaining the full set of query tokens. Link to demo video is \href{https://www.youtube.com/watch?v=e73aE7rFGrg}{\underline{here}}.}}
\label{fig:demo}
\end{figure}

In the quest for computational efficiency, the concept of sparse attention has gained traction. Methods like Token Merging (ToMe)~\cite{bolya2023tome} and its application in latent image diffusion models~\cite{tomesd} have reduced the computation time required by condensing tokens with high similarity, thereby retaining the essence of the information with fewer tokens. Similarly, approaches like Neighborhood Attention~\cite{hassani2023neighborhood} and Focal Transformers~\cite{focal} have introduced mechanisms where query tokens attend only to a select neighborhood, balancing the trade-off between receptive field and computational load. These strategies aim to efficiently approximate the attention mechanism's output. While performant, these methods typically require training-time modifications to be successful, incurring significant logistical overheads to leverage their optimizations.

Complementing the sparse attention frameworks, attention approximation methods offer an alternative avenue by exploiting mathematical properties to simplify the attention operation. Techniques ranging from replacing the softmax with more computationally friendly nonlinearities~\cite{chen2020arelu}, to fully linearizing attention~\cite{katharopoulos2020transformers}, and leveraging the kernel trick for dimensionality reduction~\cite{choromanski2022rethinking}, have been explored to approximate attention efficiently but are also generally required to be trained into the model. 

Building upon these works and aiming to address the pretraining requirement, we propose a novel post-hoc method for accelerating inference, which we refer to as Token Downsampling (ToDo). Our approach, ToDo, is inspired by the observation that adjacent pixels in images tend to exhibit similar values to their neighbors. Hence, we employ a downsampling technique to reduce tokens, akin to grid-based subsampling in image processing. Compared to prior method ToMe \cite{tomesd}, our method not only simplifies the merging process but also significantly reduces computational overhead, as it eliminates the need for exhaustive similarity calculations. In summary, our main contributions are:

\begin{itemize}
    \item A training-free method that can accelerate inference for Stable Diffusion up to 4.5x faster, beating previous methods in balancing throughput and fidelity.
    \item An in-depth analysis of attention features within the U-Net, and hypotheses on why attention can be approximated sparsely without substantially hurting fidelity.
\end{itemize}

\section{Methods}

\subsection{Background}
\paragraph{Diffusion Models for Image Generation}The diffusion model \cite{song2019generative} employs a U-Net architecture~\cite{ronneberger2015u}  with transformer-based blocks that utilize self-attention layers~\cite{rombach2021highresolution}. This setup flattens spatial dimensions into a series of tokens, which are then fed through multiple transformer blocks to predict the denoised image.  

\paragraph{Original Token Merging Scheme} In the original ToMe~\cite{bolya2023tome} framework, tokens are categorized into source (src) and destination (dst) sets. The merging process involves identifying the $r$ most similar tokens from the src set and merging them into the dst set, effectively reducing the total token count by $r$. This merging is defined as $x_{\text{merged}} = \frac{1}{r}\sum_{i=1}^{r}x_i $ where $x_i$ represents individual tokens to be merged.
\medskip

Overall, the original ToMe method is predicated on the reduction of computational load through merging of similar tokens prior to being input to attention layers. This process involves the computation of a similarity matrix, where tokens exhibiting the highest similarity are merged. Subsequently, the unmerging process aims to redistribute the merged token information back to the original token locations. This approach, however, introduces two critical bottlenecks:
\begin{itemize}
    \item \textbf{Computational Complexity:} The similarity matrix calculation, $\mathcal{O}(n^2)$ complexity, is costly in itself, especially when required at every step of the process.
    \item \textbf{Quality Degradation:} The merge-unmerge cycle inherent to ToMe can lead to significant loss of image detail, particularly at higher merging ratios.
\end{itemize}

\subsection{Training Free Enhancements} Our proposed token downsampling (ToDo) methodology extends the original ToMe approach, addressing its computational bottlenecks and quality degradation issues when applied to Stable Diffusion models. We introduce two principal modifications with ToDo:  an optimized token merging method based on spatial contiguity and a refined attention mechanism that mitigates the need for unmerging. 

\paragraph{Optimized Merging Through Spatial Contiguity} 

We introduce a novel token merging strategy that leverages the inherent spatial contiguity of image tokens. Recognizing that tokens in close spatial proximity exhibit higher similarity, thus providing a basis for merging without the extensive computation of pairwise similarities. Therefore, we employ a downsampling function $D(\cdot)$ using the Nearest-Neighbor algorithm \cite{bankman2008handbook}. We note this approach is akin to strided convolutions, as shown in Figure~\ref{fig:demo}. Formally, let $T = \{t_1, t_2 \ldots t_n \}$ denote the original set of image tokens arranged in a two-dimensional grid reflecting their spatial relationships. The proposed downsampling operation, $D$ is applied to $T$ to yield a reduced set of merged tokens $T'$, as such:
\begin{align*}
    T' = D(T) = \{D(t_1), D(t_2) \ldots D(t_m) \} \hspace{0.5em} \text{, where $m<n$}
\end{align*}

This enhancement mitigates the computational overhead associated with the pairwise similarity computation inherent in ToMe. By leveraging the assumption that spatially adjacent tokens are more likely to be similar, we bypass the need for $\mathcal{O}(n^2)$ similarity calculations, instead employing a more computationally efficient $\mathcal{O}(n)$ downsampling operation. 

\paragraph{Enhanced Attention Mechanism with Downsampling} To mitigate the information loss inherent to the unmerging process in conventional token merging approaches, we introduce a refinement to the attention mechanism within the transformer architecture \cite{vaswani2023attention}. This modification entails the application of the downsampling operation $D(\cdot)$ to the keys, $K$, and values $V$ of the attention mechanism while preserving the original queries $Q$. The modified attention function can be mathematically articulated as follows, with $d_k$ denoting the dimensionality of the keys, ensuring proper scaling within the softmax operation.

\begin{align*}
    \text{Attention}(Q,K,V) = \text{softmax}\bigg( \frac{Q\cdot D(K)^T}{\sqrt{d_k}}\bigg)\cdot D(V)
\end{align*}

This refinement ensures that the integrity of the queries is preserved, thereby maintaining the fidelity of the attention process while reducing the dimensionality of the matrices involved in the attention computation.

\section{Experiments}

% TODO:
%   * Qualitative results of generated images using our method vs tome vs flash attention (probably one or two sets are enough due to space constraints)
%   * One line graph showing average diffusion time on y-axis against a resolution sweep on the x-axis at some fixed token merging ratio
%   * One line graph showing average diffusion time on y-axis against token merging ratio on the x-axis at some fixed resolution
%   * Could also include the hbf / reconstruction scores on a second y-axis on both the above plots to avoid having to use a separate table, due to space constraints
%   * Generate figures in .tex format for highest quality rendering on paper

%------------------------------------------------------------------------- 

\begin{figure}[!htbp]
% \centering
\includegraphics[width= 0.9\columnwidth]{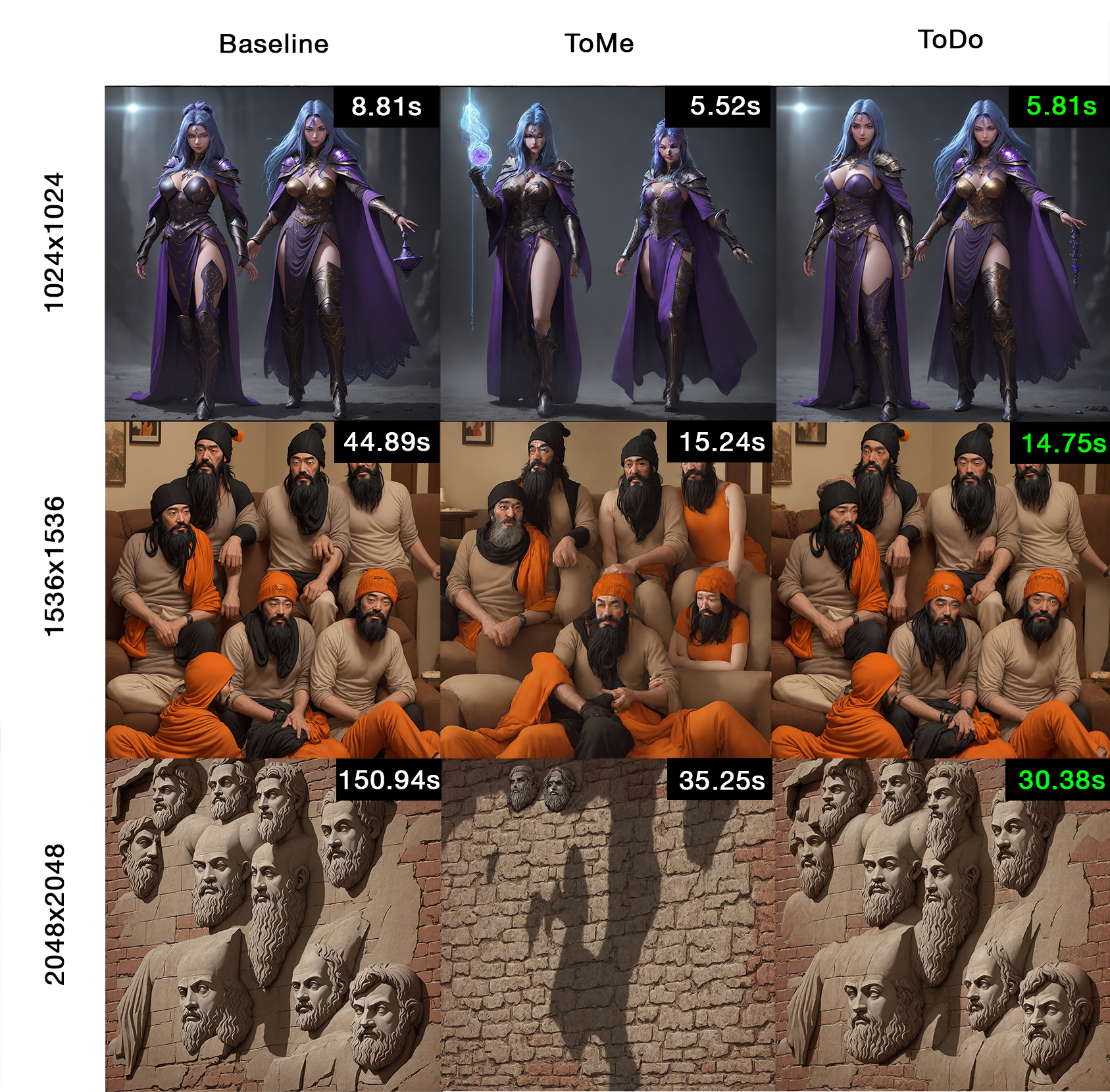}
\caption{{Qualitative comparison of attention methods with: 25\% of tokens at $1024\times 1024$, 11\% at $1536 \times 1536$, and 6\% at $2048\times 2048$, maintaining a consistent token count of 4096 post-merging.}}
  \label{fig:grid}

\end{figure}

\paragraph{Experimental Setup}
For our empirical evaluation, we employ the finetuned DreamshaperV7 model \cite{luo2023latent}, noted for its superior handling of larger image dimensions which are central to this study. All experiments are conducted on a single A6000 GPU, utilizing float16 precision and flash attention~\cite{dao2022flashattention}  for inference as this has become the norm for many users. We use the DDIM sampler~\cite{song2020denoising} with 50 diffusion steps and a guidance scale of 7.5~\cite{diffusersopt}. Each experiment involves averaging 10 generations comparing ToDo against ToMe with baseline referring to standard generations without token merging. The resolutions benchmarked include: $1024\times1024$, $1536\times1536$ and $2048\times 2048$ across two token merging ratios, 0.75 and 0.89 which denotes the proportion of tokens removed. This is equivalent to 2x and 3x downsample respectively. For the comparison in Figure \ref{fig:grid} we also use a merge ratio of 0.9375 for the $2048 \times 2048$ images, equivalent to a 4x downsample.

% \paragraph{Attention Matrix Rank Estimation} 
% To ascertain the attention mechanisms' efficiency, we examine the rank of attention matrices  $A = \text{softmax}(QK^T/\sqrt{d})$ through Singular Value Decomposition (SVD). Given the computational intensity of SVD, especially for large matrices, we employ full SVD for $512 \times 512$ image attention matrices and resort to random SVD approximation for larger matrices. This analysis is confined to two randomly selected heads out of the eight available, maintained consistently throughout the study to ensure comparability.

% \paragraph{Token Merging Evaluation} Building on the original token merging approach, we identify and address two primary limitations: increased computational complexity and quality degradation at higher merging ratios. Our strategy replaces the complex similarity calculation and merging-unmerging cycle with a streamlined downsampling operation, positing that adjacent tokens are inherently similar. We explore various downsampling techniques, including nearest-neighbor, bicubic, and bilinear interpolations, to determine the most effective method in preserving image fidelity while enhancing processing speed. Further, we refine the token merging approach by exclusively merging key and value tokens in the attention mechanism, obviating the need for unmerging and potentially preserving more image details. This modification is hypothesized to retain the performance benefits of token merging without the associated quality losses

% \subsection{Experimental Results}

\paragraph{Image Quality and Throughput}

\begin{figure}[htbp]
\centering
\includegraphics[width= 0.9\columnwidth]{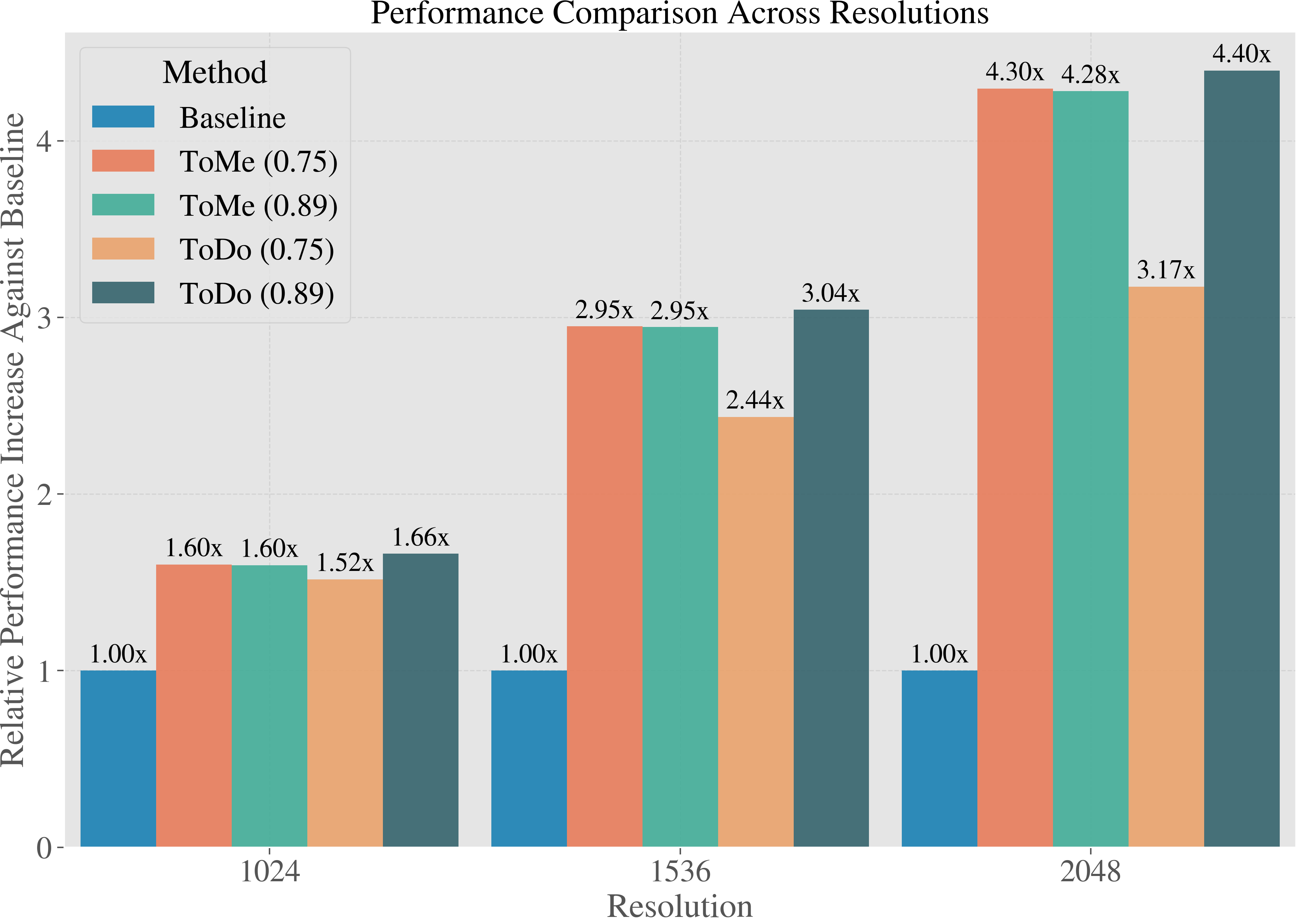}
\caption{{Inference throughput, measured in seconds, across resolutions using attention methods at various merge ratios, with bars representing the relative performance increase against the baseline.}}
  \label{fig:comp}

\end{figure}

To assess the fidelity and detail preservation of generated images, we utilized Mean Squared Error (MSE) to quantify each method's deviation from the baseline, and High Pass Filter (HPF) a standard for evaluating image sharpness and texture preservation \cite{gonzalez2009digital}. Our analysis, substantiated by Figure~\ref{fig:grid} and Table \ref{tab:results}, demonstrates that our method not only closely mirrors the baseline in terms of MSE but also maintains comparable HPF values, underscoring its efficiency in retaining image features while ensuring higher throughput, as depicted in Figure \ref{fig:comp}.

\begin{table}[!htbp]
\centering

\label{tab:results}
\resizebox{0.6\columnwidth}{!}
{
\begin{tabular}{@{}lccr@{}}
\toprule
\textbf{Method}                & \textbf{Merge Ratio} & \textbf{MSE} & \textbf{HPF} \\ \midrule
\textit{Baseline}              & -                    & -            & \textit{4.846} \\ [1.0ex] % \hline \\
\multirow{2}{*}{ToMe}          & 0.75                 & $2.686 \times 10^{-2}$     & 4.022        \\ 
                               & 0.89                 & $2.671 \times 10^{-2}$      & 4.003        \\ [1.0ex] % \hline \\
\multirow{2}{*}{ToDo (ours)}   & 0.75                 & \textbf{$6.247 \times 10^{-3}$} & \textbf{4.887} \\
                               & 0.89                 & \textbf{$9.207 \times 10^{-3}$}  & \textbf{4.733} \\ \bottomrule
\end{tabular}
}
\caption{ Metrics from various attention methods, averaged over 10 generations of different prompts at $1536\times 1536$ resolution. MSE denotes the mean squared error relative to the baseline, while HPF represents the mean absolute magnitude post-high pass filtering.}
\label{tab:results}
\end{table}
% Inserting the .tex plot

% \begin{figure}[htbp]
%     \centering
%     \resizebox{0.85\columnwidth}{!}{
%     {
%         \scriptsize
%         \input{IJCAI-Demo/images/graph_perf_text_sq}
%     }}
%     \caption{{Inference throughput across resolutions using different attention methods at various merge ratios, with bars representing the relative performance increase against the baseline .}}
%   \label{fig:comp}

% \end{figure}

% In the exploration of hidden features within the Unet, we find that neighboring tokens exhibit high similarity as predicted in addition to a few trends. First, similarity values decrease throughout the denoising process. This may be explained by the fact that diffusion models first synthesize coarse, low frequency detail in early steps and finer details in later steps \cite{yang2022diffusion}. Secondly, we observe a lack of clear trend in similarity differing across different depths. As the spatial compression of the hidden features increases it may expected that informational redundancy would decrease. We place most attention on the lowest similarity measurement within each neighborhood, as this serves as a lower bound to local informational redundancy and an upper bound for informational loss when sub-sampling.
\paragraph{Latent Feature Redundancy}

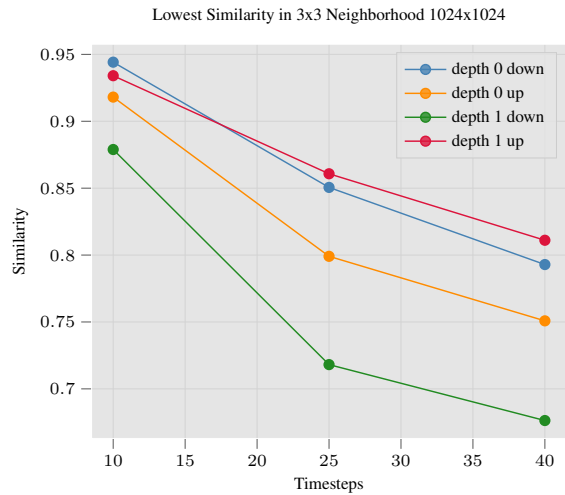
\begin{figure}[!htbp]
    \centering
    \resizebox{0.9\columnwidth}{!}{
    {
        \scriptsize
        % This file was created by tikzplotlib v0.9.8.
\begin{tikzpicture}

\definecolor{color0}{rgb}{0.274509803921569,0.509803921568627,0.705882352941177}
\definecolor{color1}{rgb}{1,0.549019607843137,0}
\definecolor{color2}{rgb}{0.133333333333333,0.545098039215686,0.133333333333333}
\definecolor{color3}{rgb}{0.862745098039216,0.0784313725490196,0.235294117647059}

\begin{axis}[
axis background/.style={fill=white!89.8039215686275!black},
axis line style={white},
legend cell align={left},
legend style={
  fill opacity=0.8,
  draw opacity=1,
  text opacity=1,
  draw=white!80!black,
  fill=white!89.8039215686275!black
},
tick align=outside,
tick pos=left,
title={Lowest Similarity in 3x3 Neighborhood 1024x1024},
x grid style={white!82.7450980392157!black},
xlabel={Timesteps},
xmajorgrids,
xmin=8.5, xmax=41.5,
xtick style={color=white!33.3333333333333!black},
y grid style={white!82.7450980392157!black},
ylabel={Similarity},
ymajorgrids,
ymin=0.66294007660905, ymax=0.957558957640234,
ytick style={color=white!33.3333333333333!black}
]
\addplot [semithick, color0, mark=*, mark size=2, mark options={solid}]
table {%
10 0.944167190320635
25 0.850567606842883
40 0.792913009894322
};
\addlegendentry{depth 0 down}
\addplot [semithick, color1, mark=*, mark size=2, mark options={solid}]
table {%
10 0.918062291903056
25 0.799066615570015
40 0.750836928225311
};
\addlegendentry{depth 0 up}
\addplot [semithick, color2, mark=*, mark size=2, mark options={solid}]
table {%
10 0.878891205533291
25 0.718125754588082
40 0.676331843928649
};
\addlegendentry{depth 1 down}
\addplot [semithick, color3, mark=*, mark size=2, mark options={solid}]
table {%
10 0.934043507180759
25 0.860763206258925
40 0.811068555889651
};
\addlegendentry{depth 1 up}
\end{axis}

\end{tikzpicture}
    }}
\caption{{Lowest cosine similarity between tokens in a $3\times3$ area across diffusion timesteps and U-Net locations  extracted from 10 generations of different prompts at $1024 \times 1024$. Timesteps out of 50 indicate noise reduction; Depth 0 is initial resolution, Depth 1 is after 2x downsampling. Up/down denotes encoder/decoder blocks.}}
    \label{fig:sdas}

\end{figure}

We investigated latent feature redundancy in the Stable Diffusion U-Net, assessing similarity among adjacent latent features. By extracting latent representations at various stages and noise levels, we constructed cosine similarity matrices, focusing on the proportion of tokens with top-3 similarities within a $3\times3$ area, and the highest, mean, and lowest similarities within $3\times3$  and $5\times5$  areas. 
We observed high similarity among neighboring tokens within the hidden features and notable trends as seen in Figure~\ref{fig:sdas}. Similarity trends varied across different depths without a distinct pattern, possibly due to the increasing spatial compression and consequent reduction in information redundancy with values diminishing as the denoising progresses, likely because diffusion models initially generate broad details and later refine them.
 % We focus on the lowest similarity in each neighborhood, indicating minimal local redundancy and maximum potential information loss during sub-sampling.

\section{Conclusion}
We demonstrate that our approach ToDo is capable of  maintaining the balance between efficient throughput and fidelity, especially in high-frequency components.We also show that nearby features within the U-Net might be redundant and postulate that our method can benefit other attention based generative image models, especially those operating on a large number of tokens. Future work can explore the differentiability of our method, and leverage it to efficiently finetune Stable Diffusion at previously unseen larger image dimensions.

% \pagebreak
% \pagebreak
% .
% \pagebreak

% .

\newpage
%% The file named.bst is a bibliography style file for BibTeX 0.99c
\bibliographystyle{named}
\bibliography{paper}

\end{document}